\documentclass{article}
\usepackage{spconf,amsmath,graphicx,hyperref}
\usepackage{multirow}
\usepackage{amsfonts}
\usepackage{graphicx}
\usepackage{subcaption}
\usepackage{graphicx}
\usepackage{caption}
\usepackage{array}
\usepackage{adjustbox}
\usepackage{algorithmic}
\usepackage[linesnumbered,ruled,vlined]{algorithm2e}

\title{Morphology-Aware KOA Classification: Integrating Graph Priors with Vision Models}
\address{Author Affiliation(s)}
\name{
\begin{tabular}{c}
Marouane Tliba$^{1,3}$, Mohamed Amine KERKOURI$^{2}$, Yassine Nasser$^{2}$, Nour Aburaed$^{5}$ \\
Aladine Chetouani$^{3}$, Ulas Bagci$^{4}$, Rachid Jennane$^{1}$
\end{tabular}
}

\address{
{$^{1}$}University of Orleans, Orleans, France; 
{$^{2}$}F-initiatives, Paris, France; \\
{$^{3}$}University of Sorbonne Paris Nord, Paris, France;
{$^{4}$}Northwestern University, Chicago, USA; \\
{$^{5}$}University of Dubai, Dubai, UAE
}
\begin{document}
%
\maketitle
\begin{abstract}
Knee osteoarthritis (KOA) diagnosis from radiographs remains challenging due to the subtle morphological details that standard deep learning models struggle to capture effectively. We propose a novel multimodal framework that combines anatomical structure with radiographic features by integrating a morphological graph representation—derived from Segment Anything Model (SAM) segmentations—with a vision encoder. Our approach enforces alignment between geometry-informed graph embeddings and radiographic features through mutual information maximization, significantly improving KOA classification accuracy. By constructing graphs from anatomical features, we introduce explicit morphological priors that mirror clinical assessment criteria, enriching the feature space and enhancing the model's inductive bias. Experiments on the Osteoarthritis Initiative dataset demonstrate that our approach surpasses single-modality baselines by up to 10\% in accuracy (reaching nearly 80\%), while outperforming existing state-of-the-art methods by 8\% in accuracy and 11\% in F1 score. These results underscore the critical importance of incorporating anatomical structure into radiographic analysis for accurate KOA severity grading. 
\end{abstract}
\section{Introduction}
\label{sec:intro}
\vspace{-3mm}
Knee osteoarthritis (KOA) is a degenerative joint disease marked by progressive cartilage loss, osteophyte formation, and subchondral bone remodeling, leading to debilitating symptoms such as chronic pain, stiffness, and swelling. Radiography remains the gold standard for diagnosing KOA due to its widespread availability, low cost, and rapid imaging capabilities. Radiographic assessment relies on the Kellgren-Lawrence (KL) grading system \cite{KL}, which classifies disease severity into five stages (0–4) based on structural biomarkers like osteophyte and joint space narrowing (JSN) width \cite{B5}.

Recently, various deep Learning approaches have emerged to diagnose knee OA using X-ray images \cite{B1,B2,B4}. 
In~\cite{Nguyen}, Nguyen et al. proposed a semi-supervised framework that leverages the mixup algorithm to synthesize out-of-distribution samples, enforcing prediction consistency and improving model robustness in scenarios with scarce labeled training data. Nasser et al. \cite{Nasser} propose a Discriminative Shape-Texture Convolutional Network that embeds a Gram Matrix Descriptor block to compute texture features from intermediate CNN layers~\cite{CNN},
which are fused with high-level shape features to improve early-stage OA detection. 
\cite{Sekhri24} employed an a hierarchical representation learning  approach that helps propagate low level features deeper into the network, thus enriching the representation ability of the network.
While deep learning has shown promise in automating KOA assessment, conventional vision models often overfit to local textural artifacts rather than relevant anatomical structures. Most of SoTA methods work directly on the raw pixel image and neglect the fact that the KL grading system is based on structural biomarkers \cite{KL,B4,B5}.

In this work, we address these challenges by redefining how structural relationships are encoded. We posit that the spatial and morphological relationships between the femoral and tibial bones provide critical diagnostic information. To capture this, we construct a graph from SAM-derived joint masks \cite{SAM}, explicitly encoding the anatomical geometry of the joint. This graph, which reflects key morphological attributes (e.g., joint space width, bone curvature), is leveraged to generate compact, discriminative embeddings that complement the deep features extracted by a vision backbone.
Motivated by the need to reduce high-dimensional pixel-level complexity and guide learning toward anatomy-aware representations, our framework integrates graph-based and vision-based modalities. By minimizing a hybrid objective function, we force our network to align these complementary feature spaces through mutual information maximization and a learnable cross-modal translation module while optimizing for accurate KOA classification.
Overall, our work aims to develop a robust and interpretable diagnostic framework that overcomes the limitations of conventional models by incorporating explicit anatomical priors derived from graph representations. These priors enrich feature learning with clinically validated biomarkers (e.g., KL-grade severity criteria), bridging the gap between data-driven learning and radiological expertise.
\vspace{-5mm}
\section{Proposed Method}
\begin{figure*}[t]
    \centering
    \includegraphics[width=0.7\textwidth]{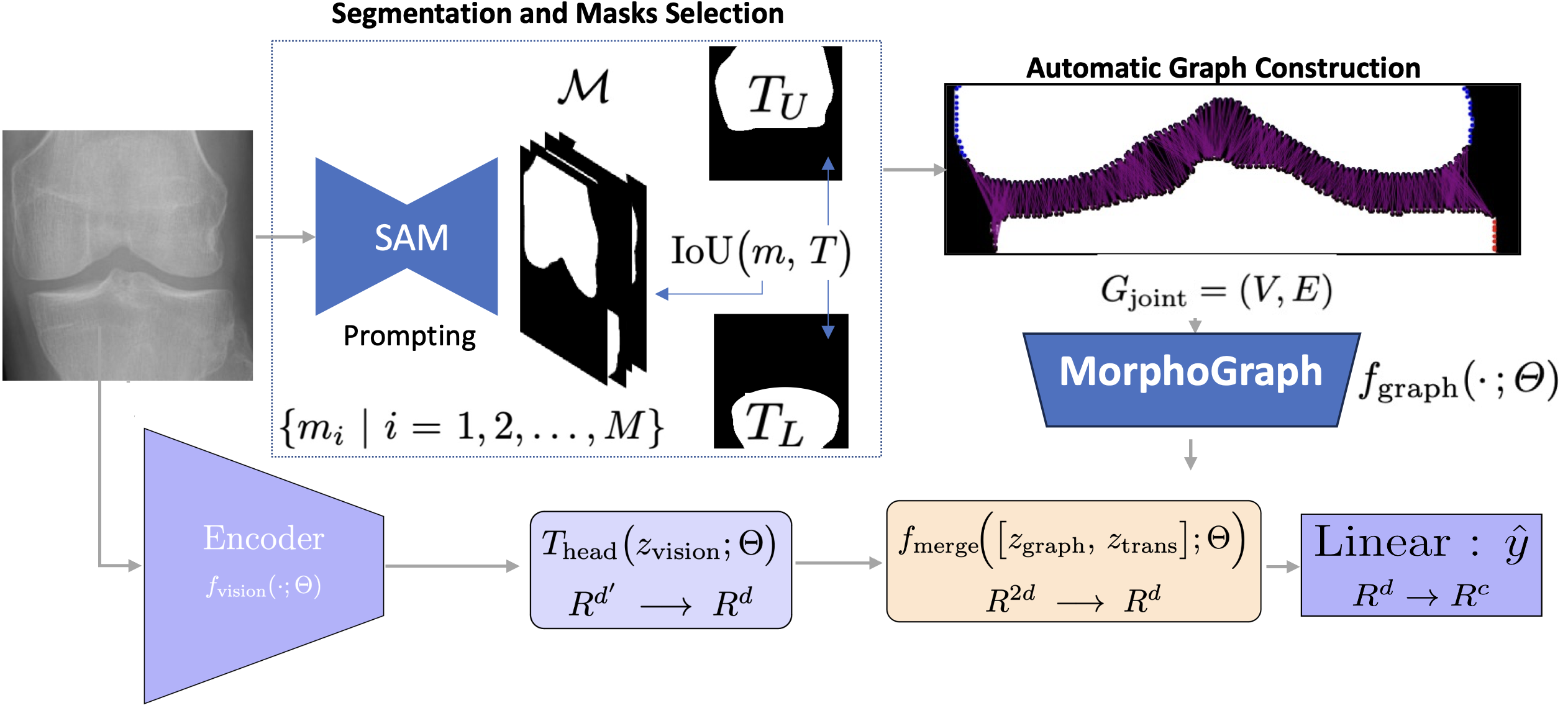}
    \vspace{-3mm}
    \caption{%
    Overview of our  pipeline: First, SAM is prompted to generate candidate masks \(\mathcal{M}\). The best mask \(m^*\) is chosen based on IoU with upper and lower bone templates \((T_U, T_L)\). Next, a morphological graph \(G_{\mathrm{joint}}\) is constructed from the joint boundary and processed by the graph encoder \(f_{\mathrm{graph}}(\cdot;\Theta)\). Simultaneously, the radiograph is passed through a vision encoder \(f_{\mathrm{vision}}(\cdot;\Theta)\). The translation head \(T_{\mathrm{head}}(z_{\mathrm{vision}};\Theta)\) aligns the vision embedding with the graph domain, and the fusion module \(f_{\mathrm{merge}}([\;z_{\mathrm{graph}},\,z_{\mathrm{trans}}];\Theta)\) combines both representations. Both \(T_{\mathrm{head}}\) and \(f_{\mathrm{merge}}\) are linear multi perceptron layers. Finally, a linear layer produces the KOA severity logits \(\hat{y}\).%
    }
    \label{fig:koa-overview}
    \vspace{-6mm}
\end{figure*}

\vspace{-3.5mm}
We propose a fully automated end-to-end multimodal approach that integrates graph-based morphological cues with radiographic features for robust KOA severity assessment. 
An overview of the full pipeline is depicted in Fig.~\ref{fig:koa-overview}.
\vspace{-3mm}
\subsection{Automatic Graph Construction}

\noindent
\textbf{Mask Selection.}\quad
Given an input radiograph \(X \in \mathbb{R}^{H \times W \times C}\), we prompt the Segment Anything Model (SAM)~\cite{SAM} using two dense point grids to generate a set of candidate segmentation masks, \(\mathcal{M} = \{ m_i \mid i = 1,2,\dots,M \}\). To isolate the joint region, we identify the optimal masks \(m^*_U\) and \(m^*_L\) for the upper and lower bones, respectively, by comparing each \(m_i\) against predefined anatomical templates \(T_U\) and \(T_L\).These templates are crafted to capture representative femoral and tibial morphologies spanning the full spectrum of KOA severity. The selected mask \(m^*\) maximizes the intersection-over-union (IoU) with its corresponding template:
\(
m^* = \arg\max_{m \in \mathcal{M}} \mathrm{IoU}(m, T).
\)

\noindent
\textbf{Graph Construction.}\quad
From the segmented joint boundary, we uniformly sample \(N\) points 
\(\{p_i\}_{i=1}^{N}\), where each \(p_i=(x_i,y_i)\in\mathbb{R}^2\) ensures equidistant coverage of the bone contour. These points form the vertex set \(V=\{p_i\}\) of an undirected graph 
\(
G_{\mathrm{joint}} = (V, E).
\)
To define the edge set \(E\), we perform a \(k\)-nearest neighbor search under a threshold \(\tau\), retaining only edges in the relevant region.
Specifically, for each \(p_i\in V\), let:
$\mathcal{N}(p_i) =
\left\{
p_j \;\middle|\; \|p_i - p_j\|_2 \leq \tau
\right\}
\cup
\kappa_k(p_i),$
where \(\kappa_k(p_i)\) returns the \(k\)-nearest neighbors of \(p_i\). The edge set is then defined as :
$\mathcal E =
\left\{
(p_i, p_j) \;\middle|\; p_j \in \mathcal{N}(p_i)
\right\}$
If this graph is initially disconnected, we iteratively increase \(\tau\) until full connectivity is achieved. The resulting \(G_{\mathrm{joint}}\) succinctly captures global joint morphology, offering a geometric prior for KOA assessment. Thus allows our approach to encode broader anatomical cues that might otherwise be neglected by local-intensity–driven models\cite{CNN,VIT}.

\vspace{-3mm}
\subsection{\textit{\textbf{MorphoGrpah}}: Graph Morphological Classification}
\label{sec:graph-arch}
Building upon the joint graph \( G_{\text{joint}} = (V, E) \), we employ a three-layer EdgeConv-style operation~\cite{graph} to capture the geometrical relationships among its vertices. EdgeConv has proven effective in learning complex structural cues in point clouds and 3D shapes, making it well-suited for modeling the morphological geometry relevant to KOA. By stacking multiple layers, our network progressively refines the node embeddings, promoting higher-order interactions that highlight clinically significant bone features.

Let 
\(\mathbf{X}^{(\ell)} \in \mathbb{R}^{|V|\times d_\ell}\)
denote the node features at layer \(\ell\). For each node \(i\) with neighbors \(\mathcal{N}(i)\), EdgeConv produces an updated feature \(\mathbf{x}_i^{(\ell+1)}\) by aggregating local subgraphs as:

\begin{equation}
\mathbf{x}_i^{(\ell+1)} =
\max_{j \in \mathcal{N}(i)} \,\phi_{\Theta}\!\left(
\left[
\mathbf{x}_i^{(\ell)},\, \mathbf{x}_j^{(\ell)} - \mathbf{x}_i^{(\ell)}
\right]
\right)
\end{equation}

where \(\bigl[\cdot,\cdot\bigr]\) denotes feature concatenation, and \(\phi_{\Theta}(\cdot)\) is an \emph{embedding layer} parameterized by learnable weights \(\Theta\). Conceptually, \(\phi_{\Theta}\) transforms both the relative difference \(\mathbf{x}_j^{(\ell)} - \mathbf{x}_i^{(\ell)}\) and the original node feature \(\mathbf{x}_i^{(\ell)}\) into a higher-level representation of local geometry. The dimensionality of \(\phi_{\Theta}(\cdot)\) increases by a factor of two with each successive graph layer, broadening the feature capacity as it proceeds through the network.

\noindent
\textbf{Graph Normalization.}\quad
Following each EdgeConv block, we apply GraphNorm to stabilize training \cite{graphnorm}. Which is variant of instance-level normalization\cite{instancenorm}. Formally, for each node feature \(\mathbf{x}_i^{(\ell+1)}\), GraphNorm is expressed as

\begin{equation}
\mathbf{g}_i^{(\ell+1)} =
\gamma \cdot 
\frac{\mathbf{x}_i^{(\ell+1)} - \mu\bigl(\mathbf{x}^{(\ell+1)}\bigr)}{
\sigma\bigl(\mathbf{x}^{(\ell+1)}\bigr) + \epsilon} + \beta
\end{equation}

where \(\mu(\cdot)\) and \(\sigma(\cdot)\) respectively compute the mean and standard deviation over node features grouped by subgraphs or batch, and \(\gamma,\beta\) are learnable parameters.

\noindent
\textbf{Output Projection.}\quad
Finally, we combine node features via global mean and max pooling to obtain a single morphology-aware vector \(\mathbf{z}\). A learnable linear layer then maps \(\mathbf{z}\) into KOA severity logits 
\(\hat{\mathbf{y}} \in \mathbb{R}^{C}\),
where \(C\) is the number of discrete KOA grades. 
\textbf{\emph{When trained, MorphoGraph simply optimizes a cross-entropy loss over these logits}}.

\vspace{-3mm}
\section{Multi-Modal Approach}
\label{sec:multimodal}
Having established a geometry-informed representation \(z_{\mathrm{graph}} \in \mathbb{R}^{d}\) from our pre-trained (and thus frozen) graph encoder \(f_{\mathrm{graph}}(\cdot)\), we now integrate this morphological prior with a learnable vision encoder \(f_{\mathrm{vision}}(\cdot)\). Our goal is to refine the vision-based embedding \(z_{\mathrm{vision}} \in \mathbb{R}^{d'}\) such that it aligns with the structural cues captured by the graph representation, producing a unified multimodal feature space conducive to KOA severity prediction.
To accomplish this alignment, we introduce a translation module 
\(
T: \mathbb{R}^{d'} \;\to\; \mathbb{R}^{d},
\)
mapping \(z_{\mathrm{vision}}\) into the same feature dimension as \(z_{\mathrm{graph}}\). Concretely,
\(
z_{\mathrm{trans}}
= T\bigl(z_{\mathrm{vision}}\bigr).
\)
We then merge these two modalities by concatenating both inputs and pass it via a fusion network \(f_{\mathrm{merge}}\), obtaining a final joint representation \(z_{\mathrm{rep}}\in \mathbb{R}^{d}\). A classifier  is then used to map  \(z_{\mathrm{rep}}\) into KOA severity logits \(\hat{y}\).To train the model under this \textbf{classical multimodal setting}, we optimize all components excluding the pre-trained \emph{MorphoGraph} encoder using cross-entropy loss.
\subsection{Mutual Information Maximization}
To effectively align the vision embedding with the geometry-based graph representation, we introduce two complementary Mutual Information Maximization (MIM) losses that jointly train the learnable vision encoder and the translation head\cite{MIM,MMI,INFONCE,masking}. This strategy ensures that the learned representation from image modality is progressively adapts toward the fixed, morphology-rich graph features space
relevant to improving overall KOA severity prediction.

\noindent
\textbf{Adaptive Masking for Graph--Image Alignment.}\quad
During training, we guide the translation head \(T(\cdot)\) via a temporary \emph{mask ratio} \(r(e) = \max(0,\,1-e/E)\), where \(e\) and \(E\) are the current and total training steps. This scalar forms a convex blend of the original \(z_{\mathrm{graph}}\) and the translated \(z_{\mathrm{trans}}\) is defined as :
$\mathcal
z_{\mathrm{combined}} =
r(e)\, z_{\mathrm{graph}} +
\bigl(1 - r(e)\bigr)\, z_{\mathrm{trans}}$
This combined embedding is used \emph{exclusively} to align $z_{\mathrm{trans}}$ with the geometry-rich space of $z_{\mathrm{graph}}$. Specifically, we minimize an MSE loss 
\(\mathcal{L}_{\mathrm{MSE}}\) over $\|\;z_{\mathrm{graph}} - z_{\mathrm{combined}}\;\|_2^2$ to bring the two embeddings closer during training. As $r(e)$ diminishes, the emphasis gradually shifts from the fixed graph embedding to the learnable $z_{\mathrm{trans}}$, ensuring that the vision representation increasingly reflects key morphological cues.

\noindent
\textbf{Contrastive Cross-Modality Learning.}\quad
To further unify the two modalities, we maximize their mutual information via an InfoNCE loss over pairs \(\{(z_{\mathrm{graph}}^i, z_{\mathrm{trans}}^i)\}_{i=1}^{N}\).  the InfoNCE objective is

\begin{equation}
\mathcal{L}_{\mathrm{InfoNCE}} =
- \frac{1}{N}
\sum_{i=1}^{N}
\log
\frac{
\exp\left(\operatorname{sim}(z_{\mathrm{trans}}^i,\, z_{\mathrm{graph}}^i) / \tau\right)
}{
\sum_{j=1}^{N}
\exp\left(\operatorname{sim}(z_{\mathrm{trans}}^i,\, z_{\mathrm{graph}}^j) / \tau\right)
}
\end{equation}

where \(\operatorname{sim}(a,b)=\frac{a^\top b}{\|a\|\|b\|}\) denotes cosine similarity and \(\tau\) is a temperature parameter.
Encouraging positive pairs to be closer than any mismatched pairs. 
By maximizing mutual information in this manner, we align the modalities in a shared space while preserving vital morphological signals from \(z_{\mathrm{graph}}\). Ultimately, this synergy equips the vision encoder \(f_{\mathrm{vision}}\) and translation module \(T\) with a geometry-aware perspective that enhances KOA grading accuracy.

\subsection{Overall Training Objective of Multimodal Approach}
\label{sec:overall-loss}

We jointly optimize two main objectives: (i) a classification loss, which pushes the vision model to extract discriminative features for KOA severity prediction that can reside only on the Image feature domain, and (ii) a mutual information maximization objective, which aligns the vision embedding with the fixed, morphology-rich graph representation. Concretely, let
\(\mathcal{L}_{\mathrm{CE}}\)
denotes the cross-entropy loss for classifying the fused embedding into the appropriate KOA grade. By directly supervising \(f_{\mathrm{vision}}\) in this manner, we encourage it to learn radiographic cues that might not be fully reflected in the joint graph representation. Simultaneously, the mutual information losses ensure that the image embeddings adapt to—and remain compatible with—the geometry-informed space of the graph embedding. This approach shows to achieve a great leap in performance, marking state-of-the-art results. We combine these terms into a single cost:

\begin{equation}
\mathcal{L}_{\mathrm{total}} =
\lambda_{\mathrm{CE}}\, \mathcal{L}_{\mathrm{CE}} +
\lambda_{\mathrm{Info}}\, \bigl(\mathcal{L}_{\mathrm{InfoNCE}} + \mathcal{L}_{\mathrm{MSE}}\bigr)
\end{equation}

where \(\lambda_{\mathrm{CE}}\) and \(\lambda_{\mathrm{Info}}\) are weighting factors, 0.8 and 0.2 respectively. The classification term prompts the vision encoder to capture radiographic details essential for accurate grading, while the alignment term drives the image features into a geometry-aware domain. As a result, the final fused representation exploits both the high-density appearance features unique to radiographs and the morphological cues inherent in the joint graph, yielding a more robust predictor of KOA severity.

\begin{figure}[ht]
    \centering
    \includegraphics[width=\linewidth]{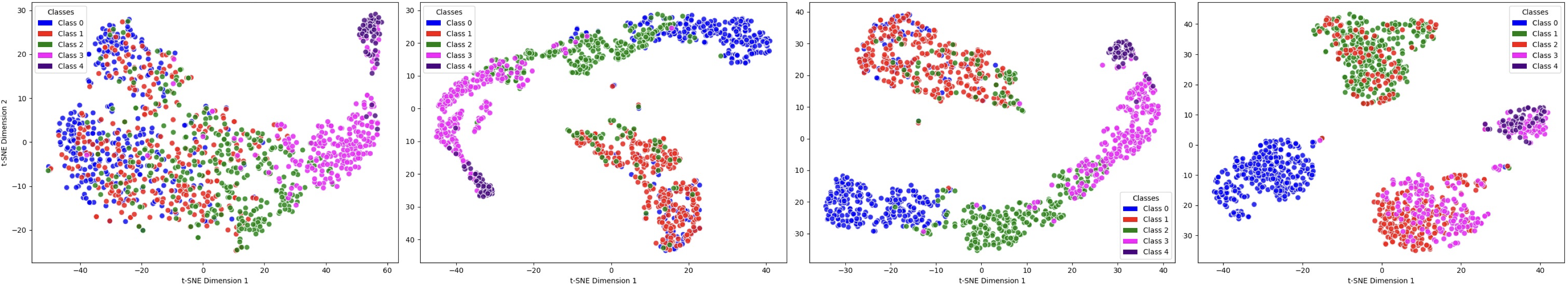}
    \vspace{-3mm}
    \caption{
    T-SNE visualizations of learned embeddings of  \textbf{MM-F (MIM) ViT Large} across four configurations. 
    From left to right: 
    (1) Vision-only model, 
    (2) classical multimodal fusion, 
    (3) multimodal fusion with mutual information maximization (MIM), 
    and 
    (4) graph-only representation.
    Colors indicate KOA severity classes.
    }
\label{fig:tsne_comparison}
\vspace{-3mm}
\end{figure}


{\small

\begin{table}
\begin{center}
\begin{tabular}{|l|c|c|c|c|c|c|}
\hline
Base  & \multicolumn{2}{c|}{Vision} & \multicolumn{2}{c|}{MM-F (Classic)} & \multicolumn{2}{c|}{MM-F (MIM)} \\
\cline{2-7}
 & Acc & F1 & Acc & F1 & Acc & F1 \\
\hline
ResNet50     & 0.657 & 0.638 & 0.745 & 0.747 & 0.776 & 0.777 \\
ResNet152    & 0.659 & 0.643 & 0.758 & 0.756 & 0.778 & 0.780 \\
ConvNeXt   & \textbf{0.696} & \textbf{0.685} & \textbf{0.768} & \textbf{0.765} & \textbf{0.794} & \textbf{0.799} \\
Swin-B       & 0.678 & 0.676 & 0.765 & 0.759 & 0.793 & 0.796 \\
ViT-Small    & 0.689 & 0.662 & 0.751 & 0.748 & 0.795 & 0.801 \\
ViT-Base     & 0.658 & 0.659 & 0.759 & 0.760 & 0.779 & 0.782 \\
ViT-Large    & 0.663 & 0.642 & \textbf{0.791} & \textbf{0.793} & \textbf{0.808} & \textbf{0.814} \\
\hline
\end{tabular}
\end{center}
\caption{Benchmarking of classification models on the OAI test set. Metrics are reported as separate Accuracy (Acc) and F1-score (F1).Vision: Vision only model , MM-F: Multimodal-Fusion, MIM: Mutual Information Maximization.}
\label{tab:mmf_results}
\vspace{-6mm}
\end{table}

}

\vspace{-6mm}
\section{Result Analysis }
\vspace{-3mm}

\noindent
\textbf{Dataset:}\quad
We evaluate our framework on the publicly available Osteoarthritis Initiative (OAI) dataset \cite{OAI}, using the baseline releases (versions 0.E.1 and 0.C.1) containing bilateral posteroanterior (PA) fixed-flexion knee X-rays. Each image is annotated with Kellgren–Lawrence (KL) grades (0–4) indicating OA severity. We extract and preprocess regions of interest (ROIs) from both left and right knees, resulting in 8,260 samples. ROIs are resized and intensity-normalized. To ensure comparability, we follow the same data splits as prior work \cite{Antony2016, Tiulpin, Wang2022} to ensure comparability : 5,778 for training, 826 for validation, and 1,656 for testing.

\subsection{Overall Performance on Radiographic Classifiers}

Table \ref{tab:mmf_results} shows the baseline performance of vision-only models on the OAI test set. Recent architectures like ConvNeXt \cite{Convnext} achieve the highest accuracy (69.63\%, F1: 68.48\%), but overall performance remains modest (65–70\%). This confirms prior findings that vision-only models struggle to capture subtle morphological cues critical for KOA grading. Thus, highlighting the limitations of relying solely on radiographic textures for KOA grading. It's worth noting that all models were trained with cross-entropy loss and pretrained weights.

\vspace{-3mm}
\subsection{Multimodal Fusion \& Mutual Information Maximization}
\label{sec:fusion-res}
Before integrating the radiographic features, we evaluate a standalone \emph{\textbf{MorphoGraph}} classifier (from Section~\ref{sec:graph-arch}). 
Then analyze the results obtained from our multi-modal approaches from Section\ref{sec:multimodal}.

\noindent
\textbf{Graph-Only Baseline:}\quad
The standalone \emph{\textbf{MorphoGraph}} classifier, leveraging graph-structured bone morphology, achieves 74.94\% accuracy and 75.19\% F1 (Table~\ref{tab:SOTAComp}), materially outperforming most vision-only models (Table~\ref{tab:mmf_results}). This result underscores the value of explicitly modeling anatomical structures and their relationships  (e.g. geometric shape of the joint, the space in the joint, ... etc.), capturing structural changes and morphological cues that are routinely overlooked by texture-centric radiographic analysis, and thereby providing a more faithful basis for KOA severity assessment.

\noindent
\textbf{Multimodal Fusion Classic(MM-F):}\quad
Supervised fusion of radiographic and morphological cues consistently delivers substantial gains across architectures, with +5–7\% accuracy improvements over vision-only baselines (Table~\ref{tab:mmf_results}). For example, ViT-Large reaches 79.11\% in MM-F Classic versus 66.30\% unimodally. These results decisively demonstrate the complementarity of texture- and morphology-based signals, yielding a more complete and clinically consonant representation of KOA severity and aligning with established rationale for KL-grade estimation

\begin{table}
\begin{center}
\begin{tabular}{|l|c|c|}
\hline
\textbf{Model} & \textbf{Accuracy} & \textbf{F1-Score} \\
\hline
Antony et al. 2016 \cite{Antony2016} & 0.5340 & 0.4300 \\
\hline
Antony et al. 2017 \cite{Antony2017} & 0.6360 & 0.5900 \\\hline
Tiulpin et al. 2018 \cite{Tiulpin} & 0.6671 & - \\\hline
Chen et al. 2019 \cite{Chen} & 0.6960 & - \\\hline
Wang et al. 2022 \cite{Wang2022} & 0.6918 & - \\\hline
Sekhri et al. 2023 \cite{B2} & 0.7017 & 0.6700 \\\hline
Sekhri et al. 2024 \cite{Sekhri24} & 0.7240 & 0.7000 \\\hline
\textbf{Ours - MorphoGraph only} & 0.7494 & 0.7519 \\\hline
\textbf{Ours - MM-F (MIM) ViT-Large} & \textbf{0.8080} & \textbf{0.8138} \\
\hline
\end{tabular}
\end{center}
\caption{State-of-the-art comparison on the OAI test set.}
\label{tab:SOTAComp}
\vspace{-6mm}
\end{table}

\noindent
\textbf{Mutual Information Maximization (MIM):}\quad
Introducing MIM as an auxiliary objective delivers a clear and material performance uplift in multimodal fusion. As reported in Table~\ref{tab:mmf_results} (MM-F MIM), the \emph{\textbf{MM-F (MIM) ViT-Large}} model attains 80.80\% accuracy and 81.38\% F1, a +1.69\% accuracy gain over MM-F Classic. MIM explicitly enforces high-fidelity alignment between vision and graph embeddings by maximizing shared information, compelling both modalities to occupy a geometry-aware latent space where texture-rich radiographic signals and anatomically grounded morphology are jointly coherent. The resulting representations are sharper and more discriminative for KOA grading. Figure~\ref{fig:tsne_comparison} provides compelling visual corroboration: the MIM-augmented model exhibits markedly tighter, class-consistent clusters with clearer inter-class margins than alternative configurations. This pronounced separation substantiates the efficacy of our multimodal framework and strengthens our central claim that integrating anatomical shape information materially improves KOA severity classification.

\vspace{-3mm}
\subsection{State-of-the-Art Comparison}

Table~\ref{tab:SOTAComp} situates our approach within the current state of the art. The \emph{\textbf{MorphoGraph-only}} model achieves 74.94\% accuracy, outperforming several prior works, including Sekhri et al. (2023) and Wang et al. (2022). When enhanced with multimodal fusion and mutual information maximization, the \emph{\textbf{MM-F (MIM) ViT-Large}} model reaches 80.80\% accuracy, an 8.16\% absolute improvement over the previous best. This substantial margin provides strong empirical validation of our central hypothesis: integrating morphological and radiographic features, coupled with advanced alignment strategies, materially advances KOA severity classification and KL-grade estimation.

\vspace{-3mm}
\subsection{Discussion and Future Work}
Our results establish that integrating geometry-driven graph embeddings with radiographic features represents a paradigm shift in KOA severity assessment. The proposed \emph{\textbf{MorphoGraph}} model, enhanced through multimodal fusion and mutual information maximization, achieves 80.80\% accuracy, surpassing prior state-of-the-art methods by over 8\%. This improvement underscores the critical role of combining anatomical structure with texture-rich radiographic cues to capture clinically relevant patterns. Future work will explore dynamic graph embeddings for modeling disease progression, self-supervised pretraining to strengthen cross-modal alignment, and the integration of clinical metadata for a more holistic and personalized KOA grading framework.

\vspace{-2mm}
\section{Conclusions}
.
We present a novel multimodal framework that integrates a geometry-centric graph representation with a learnable vision encoder for KOA classification from radiographs. By extracting precise joint masks to create anatomically informed graphs and employing mutual information maximization to align vision embeddings with morphological features, our approach significantly outperforms both single-modality baselines and existing state-of-the-art methods.
Comprehensive evaluations on the OAI dataset demonstrate substantial performance improvements, underscoring the critical importance of incorporating explicit anatomical priors in radiographic analysis. Our findings pave the way for more interpretable and robust KOA grading systems with potential clinical applications to enhance diagnostic accuracy and patient outcomes.

{

\vspace{-3mm}
\section{Acknowledgment}
\vspace{-3mm}
The authors gratefully acknowledge the support of the French National Research Agency (ANR) through the ANR-20-CE45 0013-01 project. This manuscript was prepared using data from the OAI and does not necessarily represent the views of the OAI investigators, the NIH, or private funding partners. The authors extend their sincere thanks to the study participants, clinical staff, and the coordinating center at UCSF.

\bibliographystyle{IEEEbib}
\bibliography{Template}
}

\end{document}